\documentclass[11pt]{article}
\usepackage{amssymb}
\usepackage[final]{acl}

\usepackage{times}
\usepackage{latexsym}

\usepackage[T1]{fontenc}
\usepackage[most]{tcolorbox}
\tcbuselibrary{listingsutf8}  

\definecolor{mygray}{gray}{.9}
\definecolor{c0}{cmyk}{1,0.3968,0,0.2588} 
\tcbset{
  colback=blue!5!white,
  colback=blue!3!white,
  colbacklower=blue!10!white,
  colframe=blue!40!black,
  boxrule=1.2pt,
  arc=5pt,
  left=4pt, right=4pt, top=4pt, bottom=4pt,
  fonttitle=\bfseries,
  fontupper=\ttfamily,
  enhanced,
  coltitle=white,
  drop fuzzy shadow,
  borderline={0.6pt}{0pt}{blue!55!black}
}

\usepackage[utf8]{inputenc}

\usepackage{microtype}

\usepackage{inconsolata}

\usepackage{graphicx}
\usepackage{algorithm}
\usepackage{algorithmic}
\usepackage{amsfonts}
\usepackage{float}
\usepackage{tcolorbox}
\usepackage{pifont}
\tcbuselibrary{breakable} 
\tcbuselibrary{skins} 
\usepackage{enumitem}
\usepackage{xcolor}
\usepackage{colortbl}
\usepackage{amsmath}
\definecolor{lightblue}{RGB}{200,220,240}  
\definecolor{lighterlightblue}{RGB}{235,240,252}  

\usepackage{algorithm}
\usepackage{algorithmic}
\usepackage{multirow}
\usepackage{booktabs}
\usepackage{tabularx}
\usepackage{array}
\title{GRIP: Granular Reward-Guided Parameter Interpolation \\ for Efficient Reasoning}

\newcommand{\samethanks}{\footnotemark[\value{footnote}]}

\author{
    Lam So$^{1}$\thanks{Equal contribution.},
    Canhui Wu$^{2}$\samethanks\thanks{Corresponding author.},
    Han Lin$^{1}$ \vspace{0.5em}
    \\
    $^{1}$Peking University, \hspace{0.3em}$^{2}$Xi'an Jiaotong University \\
    \texttt{wucanhui@stu.xjtu.edu.cn}
}

\begin{document}
\maketitle
\begin{abstract}
Reasoning-oriented large language models often achieve strong problem-solving performance by generating long chains of thought, but this behavior substantially increases inference cost and latency. In contrast, instruction-tuned models tend to answer more concisely, yet often lack comparable reasoning ability. This accuracy-efficiency mismatch motivates a lightweight approach that combines the strengths of both models without full model retraining. In this paper, we propose \textbf{GRIP} (\emph{Granular Reward-guided Interpolation of Parameters}), a reward-guided parameter interpolation framework for efficient reasoning. Given a reasoning model and an instruction model with identical architectures, GRIP assigns learnable interpolation ratios to individual modules and optimizes only these ratios while keeping both source models frozen. The interpolation ratios are trained with a reward signal that favors responses that are both correct and concise. Experiments show that GRIP achieves a better accuracy-efficiency trade-off than fixed or search-based merging baselines and further reveals module-wise fusion patterns associated with efficient reasoning.
\end{abstract}

\section{Introduction}

Large language models (LLMs) have demonstrated remarkable reasoning abilities across arithmetic, commonsense, and symbolic domains. Chain-of-Thought (CoT) prompting~\cite{wei2022chain,kojima2022large} further enhances these abilities by encouraging models to decompose complex problems into intermediate reasoning steps. This explicit reasoning paradigm has become an important mechanism for improving both interpretability and success rates on multi-step reasoning tasks.

However, the same explicit reasoning behavior also exposes a growing efficiency problem. Reasoning-oriented models frequently produce unnecessarily long explanations, and may even enter cycles of self-reflection when solving relatively simple problems, a phenomenon referred to as \textit{overthinking}~\cite{sui2025stop,chen2024not}. Such verbosity directly increases token consumption and latency~\cite{aytes2025sketch}, and redundant intermediate steps may introduce self-contradictions or compounding reasoning errors~\cite{cuadron2025danger,su2025between}. As reasoning models are increasingly deployed in latency- and cost-sensitive scenarios, improving reasoning efficiency without sacrificing accuracy has become a central challenge.

Existing approaches to efficient reasoning typically fall into two categories. Prompt-based methods encourage shorter responses through concise instructions or explicit token budgets~\cite{renze2024benefits,xu2025chain}, but their effectiveness depends on prompt design and may not reliably change the model's underlying reasoning behavior. Training-based methods, including reinforcement learning (RL) with length-aware rewards~\cite{luo2025o1,yi2025shorterbetter,arora2025traininglanguagemodelsreason,hou2025thinkprune} and supervised fine-tuning (SFT) on concise reasoning traces~\cite{ma2025cot,kang2025c3ot,xia2025tokenskip,yu2025long}, can more directly shape model outputs, but usually require costly model-level optimization. These limitations motivate a lighter alternative that can adjust the accuracy-efficiency trade-off without updating the full model.

Model merging provides a promising starting point for such an alternative. A reasoning model and a non-thinking instruction model from the same family often share an aligned parameter space: the former offers strong reasoning ability, whereas the latter exhibits concise instruction-following behavior. Interpolating their parameters can therefore produce fused models whose behavior moves between deliberate reasoning and concise answering. However, existing merge-based methods commonly rely on fixed global coefficients or black-box search~\cite{wu2025unlocking, wu2025revisiting}. As a result, they provide limited task-adaptive control over which modules should preserve reasoning-specific parameters and which modules can shift toward concise instruction-following behavior.

In this work, we introduce \textbf{GRIP} (\emph{Granular Reward-guided Interpolation of Parameters}), a lightweight reward-guided interpolation framework for efficient reasoning. Starting from a reasoning model and a non-thinking instruction model with identical architectures, GRIP assigns a learnable interpolation ratio to each module and optimizes only these ratios while keeping both source models frozen. At each optimization step, the current fused model rolls out responses, and a reward function jointly considers answer correctness and response length, giving higher rewards to outputs that are both correct and concise. This reward signal updates the module-wise interpolation ratios through an RL-based objective, allowing the fused model to discover task-aware fusion strategies with substantially fewer trainable parameters than full model training. Beyond improving the accuracy-efficiency trade-off over existing merging schemes, the learned interpolation patterns also provide insight into how reasoning behavior is distributed across modules.

Our contributions are summarized as follows:
\begin{itemize}
    \item We propose GRIP, a lightweight granular reward-guided parameter interpolation method that combines a reasoning model with a non-thinking instruction model of identical architecture for efficient reasoning.
    \item We optimize only module-wise interpolation ratios with an RL-based update, using rewards that jointly encourage answer correctness and response conciseness while keeping both source models frozen.
    \item Experiments demonstrate that GRIP achieves a stronger accuracy-efficiency trade-off than existing merging baselines and reveals module-wise patterns related to reasoning behavior.
\end{itemize}

\section{Related Work}

\subsection{Model Merging}
Model merging aims to integrate multiple independently trained models into a unified parameter space, enabling the combined model to inherit diverse capabilities. This paradigm has been extensively explored in settings such as continual learning~\citep{marczak2024magmax}, multi-task learning~\citep{yang2023adamerging}, and even adversarial analysis of model behaviors~\citep{gangwal2025merge}. A fundamental requirement for most merging approaches is architectural alignment, which allows parameters from different models to be directly combined.
Among existing strategies, a straightforward solution is to perform element-wise weight averaging across models~\citep{utans1996weight}. Building upon this intuition, the task arithmetic framework generalizes weight averaging by operating on task-specific parameter offsets, enabling controlled model editing and composition~\citep{ilharco2022editing}. A comprehensive overview of model merging techniques, along with their theoretical foundations and practical applications, is provided by \citet{yang2024model}.
More recently, practical deployments have demonstrated the effectiveness of model merging for reasoning efficiency. For instance, Kimi k1.5 combines models specialized in long and short chain-of-thought reasoning by uniformly averaging their parameters~\citep{team2025kimi}.

\subsection{Efficient Reasoning}

As reasoning in LLMs becomes increasingly verbose, recent works have focused on improving conciseness while maintaining reasoning quality and accuracy. Prompt-based methods, such as CCoT~\cite{renze2024benefits}, guide models through explicit instructions like ``Be concise,'' whereas CoD~\cite{xu2025chain} and Token-budget~\cite{han2024token} impose strict token constraints to prevent excessively long outputs. Supervised fine-tuning (SFT) approaches, including C3oT~\cite{kang2025c3ot}, CoT-Valve~\cite{ma2025cot}, TokenSkip~\cite{xia2025tokenskip}, and LS-Mixture~\cite{yu2025long}, train models on reasoning traces of varying lengths, with particular emphasis on shorter and more efficient chains. Reinforcement learning (RL)-based methods, such as StepPruner~\cite{wu2026beyond}, O1-pruner~\cite{luo2025o1}, ShorterBetter~\cite{yi2025shorterbetter}, and TrainEfficient~\cite{arora2025traininglanguagemodelsreason}, further encourage concise reasoning by incorporating explicit length penalties during optimization. Finally, merge-based approaches~\cite{wu2025unlocking,wu2025revisiting} reduce reasoning length by directly combining the parameters of reasoning-oriented and instruction-following models, achieving efficiency gains without additional training.

\section{Method}

\subsection{Problem Setup}

\begin{figure*}[t]
    \centering
    \includegraphics[width=\linewidth]{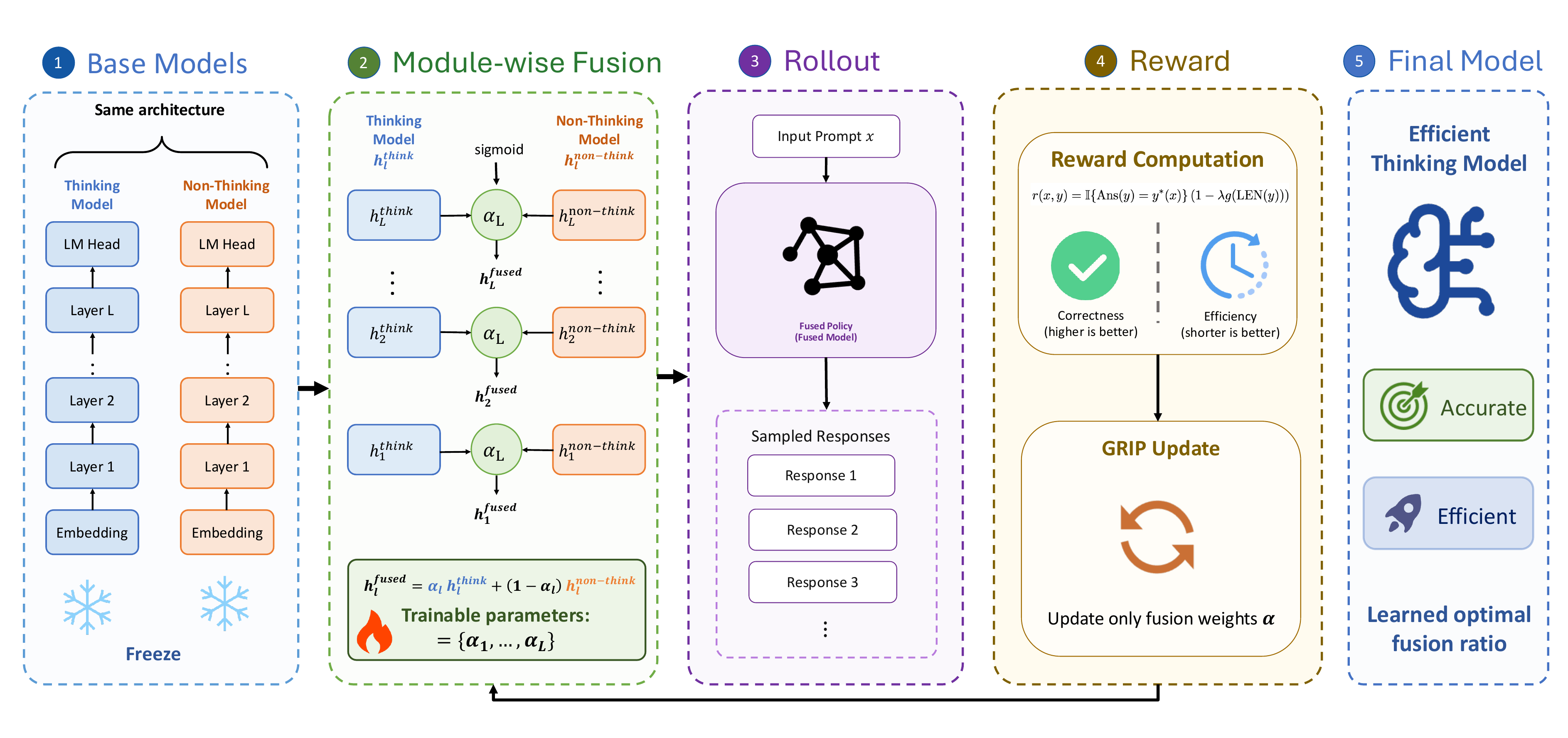}
    \caption{Overview of GRIP. GRIP learns module-wise sigmoid-controlled fusion ratios between a reasoning model and a non-thinking model, and updates them with an RL-based objective using rewards based on correctness and response length.}
    \label{fig:framework}
\end{figure*}

Let $\theta^{R}$ denote a reasoning-oriented model and $\theta^{I}$ denote an instruction-tuned model. Both models share an identical architecture and come from the same model family, so their parameter tensors are shape-compatible and directly aligned. Our goal is to construct a fused model $\theta^{F}$ that preserves the reasoning accuracy of $\theta^{R}$ while exhibiting the concise response behavior of $\theta^{I}$. GRIP parameterizes $\theta^{F}$ with a small set of module-wise interpolation logits and updates them with group-relative reward feedback in the spirit of GRPO~\cite{shao2024deepseekmath}.

Formally, we aim to optimize a set of module-wise fusion ratios such that the resulting model achieves high task accuracy with reduced output length.
Figure~\ref{fig:framework} provides an overview of the proposed interpolation and reward-guided optimization pipeline.

\subsection{Module-wise Sigmoid-controlled Fusion}

We construct the fused model by interpolating the parameters of the reasoning model and the instruction model in a module-wise manner. Let $K$ denote the number of interpolated modules, including attention modules, FFN modules, and optionally separately weighted modules such as the embedding layer and the language modeling head. When the embedding layer and the language modeling head share tied weights, they use the same interpolation coefficient to preserve weight tying. For the $k$-th module, we introduce an unconstrained trainable parameter $\rho_k \in \mathbb{R}$ and map it to a valid interpolation coefficient through a sigmoid function:
\begin{equation}
\alpha_k = \sigma(\rho_k) = \frac{1}{1+\exp(-\rho_k)}, \quad \alpha_k \in (0, 1).
\end{equation}
The fused parameters of the $k$-th module are then defined as
\begin{equation}
\theta^{F}_{k}(\boldsymbol{\rho}) =
\alpha_k \theta^{R}_{k} + (1-\alpha_k)\theta^{I}_{k},
\end{equation}
where $\alpha_k$ controls the contribution of the reasoning model and $1-\alpha_k$ controls the contribution of the instruction model. Equivalently, the overall fused model is
\begin{equation}
\theta^{F}(\boldsymbol{\rho}) =
\mathcal{F}(\theta^{R}, \theta^{I}, \boldsymbol{\alpha}), \quad
\boldsymbol{\alpha} = \sigma(\boldsymbol{\rho}).
\end{equation}
This sigmoid parameterization allows us to optimize unconstrained variables $\boldsymbol{\rho}$ with gradient-based methods while ensuring that every module-wise fusion ratio remains within the valid interval. During optimization, only $\boldsymbol{\rho}$ is updated, whereas both source models $\theta^{R}$ and $\theta^{I}$ are frozen.

Because $\theta^{F}_{k}$ is differentiable in $\rho_k$, the gradient of any per-token RL loss $\mathcal{L}$ with respect to a fusion logit follows directly from the chain rule:
\begin{equation}
\label{eq:gradient}
\frac{\partial \mathcal{L}}{\partial \rho_k}
= \left\langle
\frac{\partial \mathcal{L}}{\partial \theta^{F}_{k}},\;
\theta^{R}_{k}-\theta^{I}_{k}
\right\rangle
\cdot \sigma'(\rho_k),
\end{equation}
where $\langle\cdot,\cdot\rangle$ denotes the inner product over the module's parameter tensor. GRIP therefore receives the same per-token reward signal that full-model RL would deliver to $\theta^{F}_{k}$, but projected onto the single direction $\theta^{R}_{k}-\theta^{I}_{k}$. This projection is what makes the optimization lightweight: each module collapses to one trainable scalar without losing the per-token credit that gradient-based RL provides.

\subsection{Reward-Guided Interpolation Optimization}

We optimize the module-wise fusion logits $\boldsymbol{\rho}$ with an RL-based objective. At each optimization step, we set the current fused policy as the old policy $\pi_{\mathrm{old}}$ and sample a group of $G$ responses $\{y_i\}_{i=1}^{G}$ for each prompt $x$. Each response receives a reward that encourages correctness while penalizing excessive response length.

\paragraph{Reward Function.}
Let $y^\ast(x)$ denote the ground-truth answer of prompt $x$, let $\mathrm{Ans}(y)$ denote the final answer extracted from response $y$, and let $\mathrm{LEN}(y)$ denote the number of generated tokens in response $y$. Following the reward design in Figure~\ref{fig:framework}, we define the reward as
\begin{equation}
r(x,y)=\mathbb{I}\{\mathrm{Ans}(y)=y^\ast(x)\}
\left(1-\lambda g(\mathrm{LEN}(y))\right),
\end{equation}
where $\lambda \in [0,1]$ controls the strength of length regularization. To compare lengths among responses generated for the same prompt, we normalize response length within the \emph{correct} responses of the sampled group and apply a sigmoid soft clipping function:
\begin{equation}
g(\mathrm{LEN}(y_i)) =
\sigma\left(
\frac{\mathrm{LEN}(y_i)-\mu_x}{s_x+\delta}
\right),
\end{equation}
where $\delta$ is a small constant for numerical stability. Let $\mathcal{C}_x = \{i : \mathrm{Ans}(y_i) = y^\ast(x)\}$ denote the indices of correct responses in the group, and let $G_c = |\mathcal{C}_x|$. Then
\begin{equation}
\begin{aligned}
\mu_x &= \frac{1}{G_c}\sum_{i\in\mathcal{C}_x}\mathrm{LEN}(y_i) \\
s_x &= \sqrt{\frac{1}{G_c}\sum_{i\in\mathcal{C}_x}\left(\mathrm{LEN}(y_i)-\mu_x\right)^2}.
\end{aligned}
\end{equation}
When $G_c = 0$, the indicator $\mathbb{I}\{\mathrm{Ans}(y_i)=y^\ast(x)\}$ in the reward is zero for every response in the group, so all responses receive zero reward (and contribute zero advantage). When $G_c = 1$, $s_x$ is set to $1$ to avoid a degenerate denominator; in this case the single correct response has $z=0$ and therefore $g(\mathrm{LEN}(y_i))=\sigma(0)=0.5$.
Thus, incorrect responses receive zero reward, while correct responses are further ranked by length, with shorter correct responses receiving larger rewards.

\paragraph{Policy Update.}
Given rewards $\{r_i\}_{i=1}^{G}$ for responses sampled from the old fused policy $\pi_{\mathrm{old}}$, we compute group-relative advantages by normalizing rewards within each group:
\begin{equation}
\hat{A}_i =
\frac{r_i-\bar{r}}{\mathrm{std}(\{r_j\}_{j=1}^{G})+\delta},
\quad
\bar{r}=\frac{1}{G}\sum_{j=1}^{G}r_j.
\end{equation}
For brevity, denote the current fused policy by $\pi_{\rho}=\pi_{\theta^F(\boldsymbol{\rho})}$. Following GRPO~\cite{shao2024deepseekmath} with the \emph{clip-higher} and KL-free modifications from DAPO~\cite{yu2026dapo}, we update the fusion logits $\boldsymbol{\rho}$ using
\begin{equation}
\label{eq:objective}
\mathcal{J}(\boldsymbol{\rho})
=
\mathbb{E}\!\left[
\frac{1}{G}\sum_{i=1}^{G}
\min\!\big(
\omega_i\hat{A}_i,\;
\tilde{\omega}_i\hat{A}_i
\big)
\right],
\end{equation}
where the expectation is over prompts and response groups sampled from $\pi_{\mathrm{old}}$, $\omega_i=\omega_i(\boldsymbol{\rho})=\pi_{\rho}(y_i\mid x)/\pi_{\mathrm{old}}(y_i\mid x)$ is the policy ratio, and $\tilde{\omega}_i=\mathrm{clip}(\omega_i,\,1-\varepsilon_{\mathrm{lo}},\,1+\varepsilon_{\mathrm{hi}})$ uses asymmetric \emph{clip-higher} bounds. Both $\omega_i$ and the loss are estimated at the token level. We omit the KL anchor: since $\boldsymbol{\rho}$ is confined to a low-dimensional sigmoid-bounded space and the original model weights are frozen, the fused policy stays close to its initialization without requiring an explicit reference-policy regularizer. The trainable logits are then updated by gradient ascent:
\begin{equation}
\boldsymbol{\rho} \leftarrow
\boldsymbol{\rho} + \eta \nabla_{\boldsymbol{\rho}}
\mathcal{J}(\boldsymbol{\rho}),
\quad
\boldsymbol{\alpha}=\sigma(\boldsymbol{\rho}).
\end{equation}
This update directly changes the module-wise interpolation ratios through gradient ascent while keeping the original model parameters fixed. The overall optimization procedure is summarized in Algorithm~\ref{alg:rl_fusion}.

\begin{algorithm}[t]
\caption{RL-based Module-wise Parameter Interpolation}
\label{alg:rl_fusion}
\begin{algorithmic}[1]
\REQUIRE Reasoning model $\theta^{R}$, Instruction model $\theta^{I}$
\ENSURE Optimized fusion ratios $\boldsymbol{\alpha}^\ast$

\STATE Initialize trainable fusion logits $\boldsymbol{\rho}=\operatorname{logit}(0.8)\mathbf{1}$
\WHILE{not converged}
    \STATE Compute fusion ratios $\boldsymbol{\alpha}\leftarrow\sigma(\boldsymbol{\rho})$
    \STATE Construct fused model $\theta^F(\boldsymbol{\rho}) \leftarrow \mathcal{F}(\theta^R, \theta^I, \boldsymbol{\alpha})$
    \STATE Sample a batch of problems $\mathcal{B}$
    \STATE Set $\pi_{\mathrm{old}}\leftarrow\pi_{\theta^F(\boldsymbol{\rho})}$ and roll out response groups $\{y_i\}_{i=1}^{G}$
    \STATE Compute rewards $\{r_i\}$ using correctness and normalized length penalty
    \STATE Compute group-relative advantages $\{\hat{A}_i\}$
    \STATE Update $\boldsymbol{\rho}$ with the RL-based objective
\ENDWHILE
\STATE \textbf{return} $\boldsymbol{\alpha}^\ast=\sigma(\boldsymbol{\rho}^\ast)$
\end{algorithmic}
\end{algorithm}

\section{Experiments}
 
\subsection{Experimental Setup}
\paragraph{Models and Datasets}
We conduct experiments on the Qwen3-4B-Instruct-2507 and Qwen3-4B-Thinking-2507~\cite{yang2025qwen3} models, which share the same architecture and therefore allow direct parameter interpolation. We use DeepScaleR-preview~\cite{luo2025deepscaler} as the training dataset for optimizing the module-wise fusion logits. To test both in-domain and out-of-domain generalization, we evaluate GRIP on five reasoning benchmarks: AIME25, MATH500~\cite{lightman2023let}, GSM8K~\cite{cobbe2021gsm8k}, GPQA-D~\cite{rein2024gpqa}, and LiveCodeBench (LCB)~\cite{jain2025livecodebench}. These datasets cover competition math, grade-school math, scientific question answering, and code generation.

\paragraph{Implementation Details}
We train GRIP with the slime framework~\cite{slime_github}. For Qwen3-4B, the trainable variables are $K=74$ fusion logits: 36 for attention, 36 for FFN, one for the final RMSNorm, and one shared by the tied input embedding and LM head. This keeps optimization lightweight because both source models remain frozen and only scalar fusion parameters are updated. We use learning rate $0.1$, length penalty $\lambda=0.4$, rollout group size $G=16$, and $32$ prompts per rollout, producing $32{\times}16{=}512$ responses per rollout. With a global batch size of $512$, each rollout yields one optimizer step. The maximum response length is $10{,}240$ during training, and we train for $750$ optimizer steps. The asymmetric clipping bounds are $\varepsilon_{\mathrm{lo}}=0.2$ and $\varepsilon_{\mathrm{hi}}=0.28$; KL anchoring and entropy bonus are disabled. Evaluation uses LightEval~\cite{lighteval} with temperature $0.8$, top\_p $0.9$, and a $32{,}768$ token limit.
 
\paragraph{Baselines}  

We compare our method against three types of baselines.
1) \textbf{Qwen3 Modes}: We report the original Qwen3-Thinking and Qwen3-Instruct models to show the accuracy-efficiency trade-off before fusion.
2) \textbf{Model Merging}: We compare with representative parameter-space merging methods, including Linear interpolation, SLERP~\cite{shoemake1985animating}, TIES~\cite{yadav2023ties}, DARE-TIES~\cite{yu2024language,yadav2023ties}, and DELLA~\cite{deep2024della}. Following the setting of \citet{wu2025revisiting}, all interpolation-based baselines use a fixed reasoning-model coefficient of 0.8.
3) \textbf{Black-box Search}: We include CMA-ES~\cite{hansen2001completely} as a search-based baseline for optimizing fusion ratios without gradient-based reward feedback. To make the comparison apples-to-apples, CMA-ES uses the same training prompts as GRIP (DeepScaleR-preview), the same module-wise parameterization, and a fitness that jointly rewards accuracy and penalizes generation length on those training prompts, matching GRIP's reward function. AIME25, MATH500, GSM8K, GPQA-D, and LCB are all held-out for both methods.

\subsection{Main Results}

\begin{table*}[tb]
 \centering
 \resizebox{\linewidth}{!}{%
 \begin{tabular}{@{}lrrrrrrrrrrrr@{}}
 \toprule
 \multirow{2}{*}{\textbf{Methods}}
 & \multicolumn{2}{c}{\textbf{AIME25}}
 & \multicolumn{2}{c}{\textbf{MATH500}}
 & \multicolumn{2}{c}{\textbf{GSM8K}}
 & \multicolumn{2}{c}{\textbf{GPQA-D}$^{\dagger}$}
 & \multicolumn{2}{c}{\textbf{LCB}$^{\dagger}$}
 & \multicolumn{2}{c}{\textbf{Avg}} \\
 \cmidrule(lr){2-3}
 \cmidrule(lr){4-5}
 \cmidrule(lr){6-7}
 \cmidrule(lr){8-9}
 \cmidrule(lr){10-11}
 \cmidrule(lr){12-13}
 &Acc.&Tok.
 &Acc.&Tok.
 &Acc.&Tok.
 &Acc.&Tok.
 &Acc.&Tok.
 &Acc.&Tok. \\
 \midrule

 \rowcolor{lightgray!30}
 \multicolumn{13}{c}{\textbf{\textit{Qwen3}}} \\
 \midrule

\texttt{Thinking} 
  & 73.3 & 19630 
  & 89.4 & 5802 
  & 94.5 & 1350 
  & 66.7 & 9057 
  & 56.0 & 18439 
  & 76.0$_{(+0.0)}$ & 10856$_{(100.0\%)}$ \\ 
 
\texttt{Instruct} 
  & 36.7 & 10555 
  & 85.6 & 1623 
  & 92.6 & 304 
  & 46.5 & 459 
  & 30.3 & 2824 
  & 58.3$_{(-17.7)}$ & 3153$_{(29.0\%)}$ \\ 
 
\midrule 
\texttt{Linear} 
  & 73.3 & 15977 
  & 88.0 & 4339 
  & 93.6 & 1223 
  & 70.3 & 8591 
  & 53.1 & 15814 
  & 75.7$_{(-0.3)}$ & 9189$_{(84.6\%)}$ \\

\texttt{SLERP} 
  & 76.7 & 16467 
  & 89.0 & 4442 
  & 94.2 & 1231 
  & 69.2 & 8449 
  & 53.7 & 15770 
  & 76.5$_{(+0.5)}$ & 9272$_{(85.4\%)}$ \\

\texttt{TIES} 
  & 76.7 & 16106 
  & 85.8 & 4433 
  & 93.6 & 1197 
  & 69.2 & 8724 
  & 56.0 & 16097 
  & 76.3$_{(+0.3)}$ & 9311$_{(85.8\%)}$ \\

\texttt{DARE-TIES} 
  & 70.0 & 16455 
  & 88.0 & 4640 
  & 93.1 & 1223 
  & 68.2 & 8735 
  & 54.9 & 16658 
  & 74.8$_{(-1.2)}$ & 9542$_{(87.9\%)}$ \\

\texttt{DELLA} 
  & 66.7 & 17447 
  & 88.0 & 4807 
  & 94.5 & 1205 
  & 68.7 & 8026 
  & 53.1 & 16875 
  & 74.2$_{(-1.8)}$ & 9672$_{(89.1\%)}$ \\
  
\texttt{CMA-ES}                 
  & 60.0 & 11990                      
  & 85.8 & 3632                       
  & 94.5 & 1202            
  & 67.2 & 6839                       
  & 42.9 & 18675                                                                                             
  & 70.1$_{(-5.9)}$ & 8413$_{(77.5\%)}$ \\
 
\midrule 
\rowcolor{blue!5} 
\textbf{GRIP} 
  & \textbf{80.0} & \textbf{11838 }
  & 86.7 & \textbf{3565 }
  & 94.4 & \textbf{1115 }
  & \textbf{70.2} & 8236
  & 51.4 & \textbf{14894} 
  & \textbf{76.5}$_{(+0.5)}$ & \textbf{7930}$_{(73.0\%)}$ \\





 \bottomrule
 \end{tabular}
 }
\caption{Main comparison of accuracy and inference efficiency on five reasoning benchmarks. Acc. denotes task accuracy, and Tok. denotes the average number of generated tokens per example. The Avg column reports the mean performance across AIME25, MATH500, GSM8K, GPQA-D, and LCB. Best results are highlighted in bold. $^{\dagger}$ marks out-of-domain benchmarks: GRIP and the CMA-ES baseline are both trained on math data only (DeepScaleR-preview).}
 \label{tab:main}
 \end{table*}
 
Table~\ref{tab:main} shows that GRIP improves the accuracy-efficiency trade-off without simply moving the model toward shorter but weaker responses. Although the instruction model is much more concise, it suffers a large accuracy drop, especially on AIME25, GPQA-D, and LCB. By contrast, GRIP reduces the average generation length by 27.0\% relative to Qwen3-Thinking while slightly improving average accuracy from 76.0 to 76.5, suggesting that much of the verbose reasoning can be removed without sacrificing task-critical reasoning behavior.

GRIP also compares favorably with conventional merging baselines. Fixed-ratio methods such as Linear interpolation and SLERP shorten outputs to some extent, but their global fusion strength limits the ability to preserve reasoning-sensitive components. GRIP reaches the same average accuracy as SLERP while using 14.5\% fewer tokens, indicating that adaptive module-wise fusion provides a more efficient allocation of inference computation.

The gains vary across benchmarks. On AIME25, GRIP improves accuracy over Qwen3-Thinking by 6.7 points while reducing output length by 39.7\%, suggesting that difficult mathematical problems contain substantial redundant exploration. Notably, although GRIP is optimized only on mathematical data, it also transfers well to out-of-domain benchmarks: GPQA-D shows accuracy gains with shorter reasoning, and LCB achieves the lowest token usage among strong reasoning variants. On MATH500 and GSM8K, GRIP mainly converts the strong baseline performance into shorter generations, while LCB remains more sensitive to length reduction, possibly because code generation benefits more from extended deliberation or verification.

\subsection{Validating Module-wise Interpolation}
\label{sec:module_sweeps}

We first test whether attention and FFN require separate fusion ratios. On the Qwen3-4B Thinking/Instruct pair, we sweep $\alpha\in\{0.0,0.1,\dots,1.0\}$, where $\alpha\!=\!1$ recovers Thinking and $\alpha\!=\!0$ recovers Instruct. We compare three settings: applying $\alpha$ only to attention, only to FFN, or globally to all modules. Non-swept modules are fixed at $0.5$, isolating the marginal effect of each module type. Each setting is evaluated on the same five benchmarks (macro pass@1, average generation length).

\begin{figure*}[t]
    \centering
    \includegraphics[width=\linewidth]{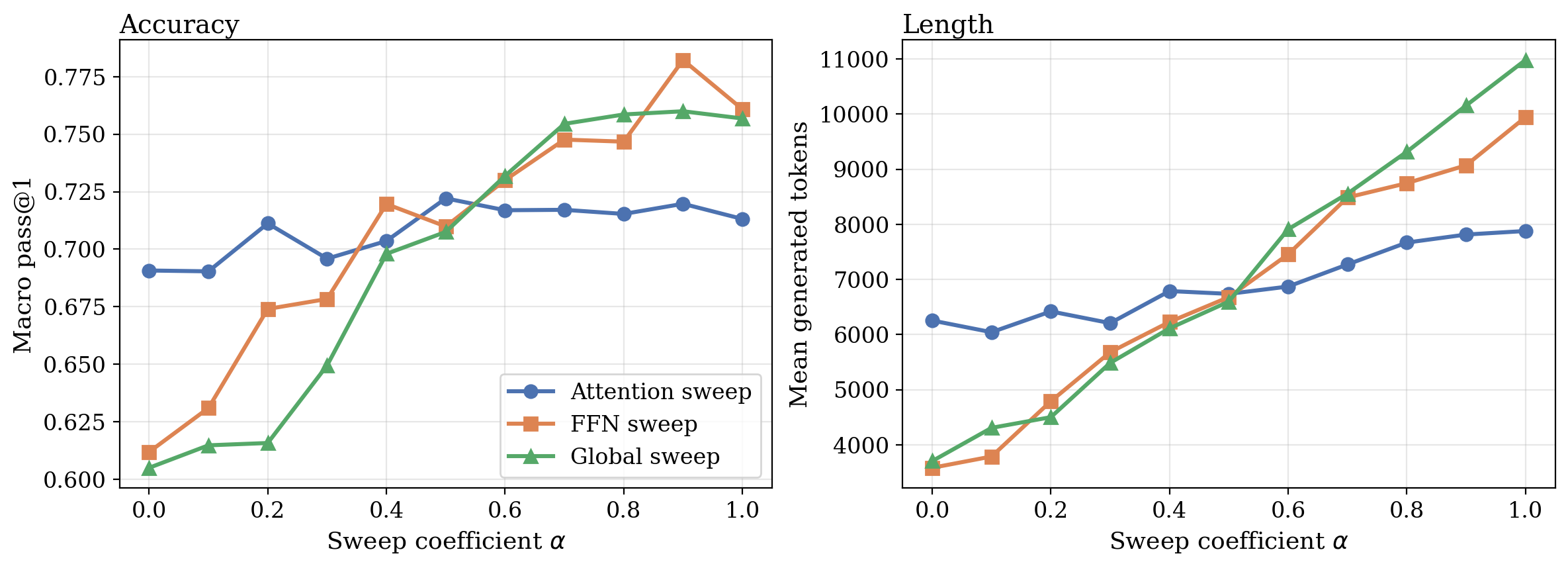}
    \caption{Module-targeted coefficient sweeps on Qwen3-4B. The attention sweep is nearly flat in both macro pass@1 and length, while the FFN sweep changes both metrics substantially. The global sweep follows the FFN trend but pays a larger length cost because all modules are moved together.}
    \label{fig:sweep_attn_ffn}
\end{figure*}

\paragraph{Attention and FFN play sharply different roles.}
Figure~\ref{fig:sweep_attn_ffn} shows a clear asymmetry. Sweeping attention barely changes macro pass@1, which remains in $[0.690,0.722]$, and increases length by only $26\%$. Sweeping FFN instead raises macro pass@1 from $0.612$ to $0.782$ and increases length by $178\%$. FFN therefore drives most of both accuracy and length, while attention is largely inert; a single global coefficient conflates the two.

\paragraph{Module-wise tuning beats any global $\alpha$.}
The FFN sweep also outperforms the global sweep for every $\alpha\geq 0.4$. Its best accuracy is $0.782$ at $\alpha\!=\!0.9$, compared with $0.760$ for the best global setting. The global sweep additionally pays an unnecessary length cost because it moves attention together with FFN. Separating the two streams removes this coupling, and motivates the per-layer extension used by GRIP.

\subsection{Tracking Module-wise Fusion Coefficients}
\label{sec:layer_dynamics}

We next track the learned fusion ratios over training. Every $\sim 10$ optimization steps, we record $\alpha_k(t)=\sigma(\rho_k(t))$ for all Qwen3-4B modules: 36 attention coefficients, 36 FFN coefficients, one shared coefficient for the tied input embedding and LM head, and one coefficient for the final RMSNorm. All coefficients start from an $80\%$/$20\%$ Thinking/Instruct blend. This lets us observe whether the optimizer keeps a nearly global interpolation pattern or actively assigns different modules to different source models.

\begin{figure}[t]
    \centering
    \includegraphics[width=\linewidth]{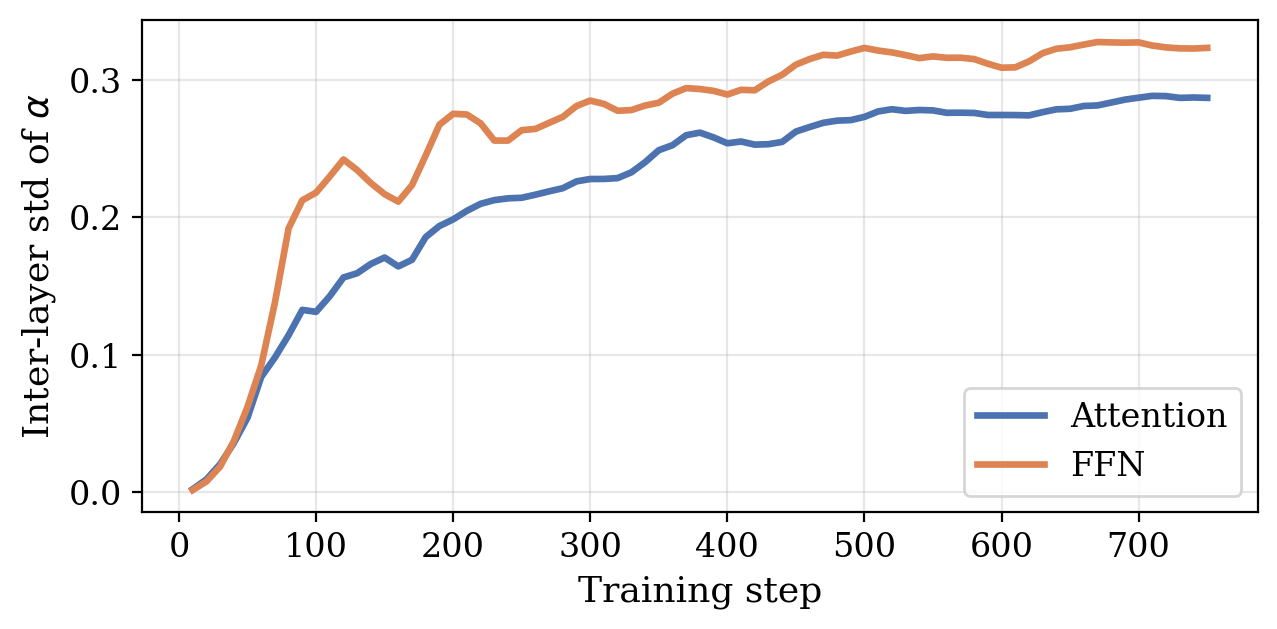}
    \caption{Inter-layer standard deviation of attention and FFN fusion ratios during training. Both start near zero, grow rapidly in the first few hundred steps, and stabilize around $\approx 0.30$, showing that training breaks the initially uniform blend.}
    \label{fig:variance_growth}
\end{figure}

\paragraph{From a uniform initialization to strong per-layer differentiation.}
Figure~\ref{fig:variance_growth} reports the inter-layer standard deviation of $\boldsymbol{\alpha}$ for attention and FFN. Both start near $0$, rise rapidly in the first $\sim 300$ steps, and saturate around $0.30$. Since a global coefficient would keep this value at $0$, the learned solution separates layers in both module streams rather than merely shifting the whole model toward one endpoint.

\begin{figure}[t]
    \centering
    \includegraphics[width=\linewidth]{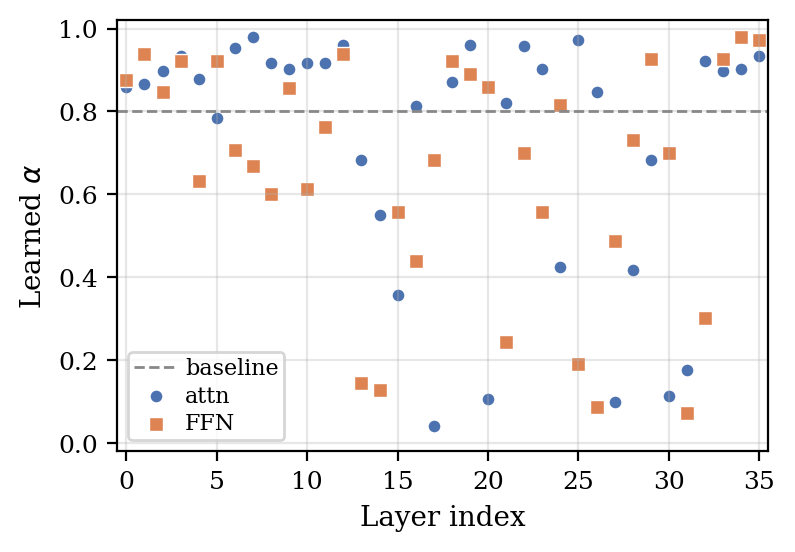}
    \caption{Learned per-layer fusion ratios after $750$ training steps for attention and FFN modules.}
    \label{fig:alpha_distribution}
\end{figure}

\begin{figure*}[t]
    \centering
    \includegraphics[width=\linewidth]{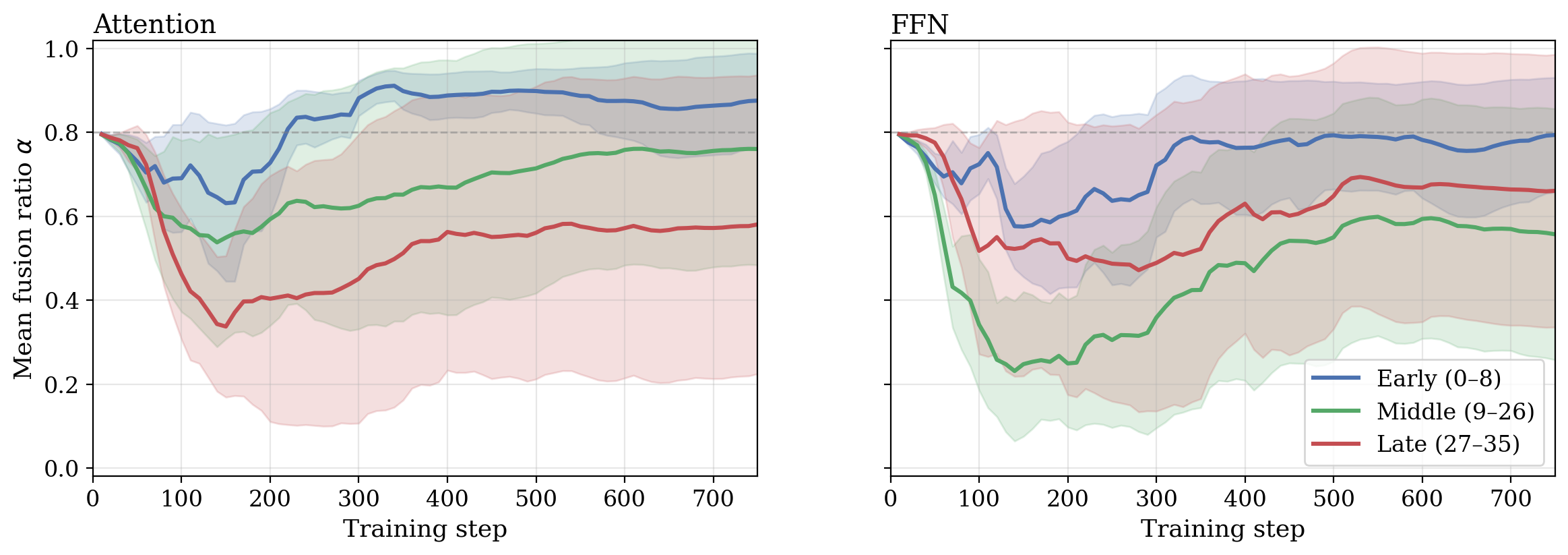}
    \caption{Mean fusion ratio $\alpha$ over training, with the $36$ layers grouped into early ($0$--$8$), middle ($9$--$26$), and late ($27$--$35$) bands; shaded regions show within-band standard deviation. Left: attention. Right: FFN. The early band stays close to the $0.8$ initialization, while middle and late bands move substantially toward Instruct, and the FFN middle band moves more aggressively than any attention band.}
    \label{fig:layer_groups}
\end{figure*}

\paragraph{Where in the network does the differentiation happen?}
Figure~\ref{fig:layer_groups} groups the $36$ layers into early ($0$--$8$), middle ($9$--$26$), and late ($27$--$35$) bands, and reports the mean fusion ratio per band over training. The split is not uniform across depth. For attention, the early band stays Thinking-heavy throughout ($\alpha\!\approx\!0.85$--$0.9$), while the middle and late bands first dip sharply toward Instruct (down to $\alpha\!\approx\!0.5$ and $\alpha\!\approx\!0.3$ respectively in the first $\sim 200$ steps) before partially recovering and stabilizing around $\alpha\!\approx\!0.7$ and $\alpha\!\approx\!0.6$. FFN follows a different shape: the early band drifts only mildly downward to $\alpha\!\approx\!0.75$, but both the middle and late bands plunge much deeper (the middle band reaches $\alpha\!\approx\!0.3$, more Instruct-heavy than any attention band) and settle around $\alpha\!\approx\!0.55$ and $\alpha\!\approx\!0.65$. Along the time axis the differentiation stabilizes rather than diverging: most movement happens in the first $\sim 300$ steps and the trajectories then flatten. Attention and FFN therefore differentiate in different depth regions and along different trajectories, which a single global ratio cannot capture. Figure~\ref{fig:alpha_distribution} compares the learned coefficients under the trained GRIP policy against the fixed $\alpha{=}0.8$ baseline used by interpolation methods following \citet{wu2025revisiting}; the broad spread around this dashed line shows that most layers prefer substantially different mixing ratios, rather than the same global coefficient.

\subsection{Reward-Guided vs.\ Black-Box Search}
\label{sec:rl_vs_cma}

We also compare GRIP with CMA-ES~\cite{hansen2001completely}, a black-box search baseline over the same module-wise parameterization $\boldsymbol{\lambda}\in[0,1]^D$ with $D{=}74$ for Qwen3-4B (one mixing weight per attn/FFN block plus embedding and final norm). CMA-ES samples $6$ candidates per generation and ranks them by the same accuracy$-$length fitness used by GRIP's reward, evaluated on the same DeepScaleR-preview prompts. With training distribution, search space, and objective all matched, the comparison isolates the optimization mechanism itself: reward-guided gradient updates vs.\ evolutionary search.

\begin{figure}[t]
\centering
\includegraphics[width=\linewidth]{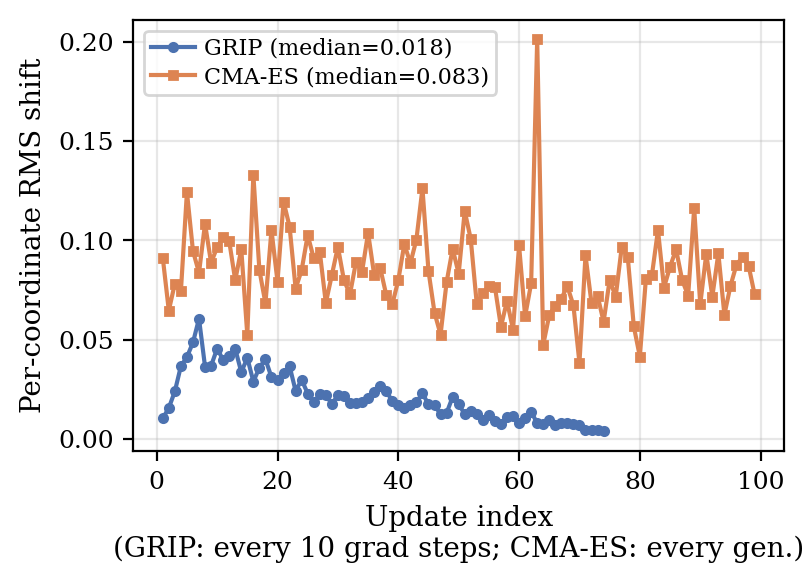}
\caption{RL optimization is smoother than CMA-ES. Per-coordinate RMS shift between adjacent updates, measured every $10$ gradient steps for RL and every generation for CMA-ES.}
\label{fig:rl_vs_cma}
\end{figure}

\paragraph{Continuity.} The median per-coordinate RMS shift between adjacent updates is $0.018$ for GRIP and $0.083$ for CMA-ES. The path-to-net-displacement ratio is $4.9\times$ for GRIP and $21.8\times$ for CMA-ES, after which CMA-ES retraces most of its path while GRIP does not.

\paragraph{Credit assignment.} CMA-ES performs $100\times6=600$ fitness evaluations on the training prompts and receives one scalar score per candidate for all $74$ coordinates, so it must infer which coordinates matter only through population-level ranking. GRIP operates on the same $74$-dim space but receives gradient feedback on every rollout, assigning credit directly to each module-level coefficient. With training data, search space, and objective held fixed, the gap in Table~\ref{tab:main} isolates the value of reward-guided updates over evolutionary search.


\section{Conclusion}

We presented GRIP, a reward-guided parameter interpolation framework for efficient reasoning. Given a reasoning model and an instruction model with identical architectures, GRIP freezes source models, assigns learnable fusion coefficients to modules, and updates these coefficients with a reward favoring correct, concise responses. Across five reasoning benchmarks, this module-wise interpolation reduces generation length while preserving or improving average accuracy relative to the original reasoning model, yielding a stronger accuracy-efficiency trade-off than fixed-ratio merging and search-based baselines. Our analyses show that reasoning relies on per-layer, per-module coefficients that cannot be reduced to a single global ratio, and that reward-guided updates yield a smoother trajectory than black-box search over the same interpolation space, with distinct fusion patterns emerging across attention and FFN modules. Together, these results suggest that reward-guided interpolation is a lightweight alternative to full-model training for reducing overthinking in large language models.

\clearpage
\section{Limitations}\label{sec:limitations}

Our study has several limitations. First, due to compute constraints we evaluate GRIP only at the $4$B scale; whether the observed accuracy-efficiency trade-off and per-layer differentiation patterns transfer to substantially larger reasoning models ($30$B+) remains open. Second, all experiments use dense Transformer backbones; we have not validated GRIP on Mixture-of-Experts architectures, where router and expert weights introduce additional structure that the current per-module parameterization does not address. Finally, GRIP assumes the reasoning model and the instruction model share an identical architecture (same depth, width, and head count), which restricts the method to within-family pairs (e.g., Qwen3-Thinking with Qwen3-Instruct); cross-family fusion (e.g., a Llama Thinking variant with a Qwen Instruct variant) is not directly supported and is not evaluated.


\bibliography{main}

@article{luo2025o1,
  title={O1-Pruner: Length-Harmonizing Fine-Tuning for O1-Like Reasoning Pruning},
  author={Luo, Haotian and Shen, Li and He, Haiying and Wang, Yibo and Liu, Shiwei and Li, Wei and Tan, Naiqiang and Cao, Xiaochun and Tao, Dacheng},
  journal={arXiv preprint arXiv:2501.12570},
  year={2025}
}

@article{yi2025shorterbetter,
  title={Shorterbetter: Guiding reasoning models to find optimal inference length for efficient reasoning},
  author={Yi, Jingyang and Wang, Jiazheng and Li, Sida},
  journal={arXiv preprint arXiv:2504.21370},
  year={2025}
}

@misc{arora2025traininglanguagemodelsreason,
    title={Training Language Models to Reason Efficiently}, 
    author={Daman Arora and Andrea Zanette},
    year={2025},
    eprint={2502.04463},
    archivePrefix={arXiv},
    primaryClass={cs.LG},
    url={https://arxiv.org/abs/2502.04463}, 
}

@article{xu2025chain,
  title={Chain of draft: Thinking faster by writing less},
  author={Xu, Silei and Xie, Wenhao and Zhao, Lingxiao and He, Pengcheng},
  journal={arXiv preprint arXiv:2502.18600},
  year={2025}
}

@inproceedings{renze2024benefits,
  title={The benefits of a concise chain of thought on problem-solving in large language models},
  author={Renze, Matthew and Guven, Erhan},
  booktitle={2024 2nd International Conference on Foundation and Large Language Models (FLLM)},
  pages={476--483},
  year={2024},
  organization={IEEE}
}

@article{ma2025cot,
  title={CoT-Valve: Length-Compressible Chain-of-Thought Tuning},
  author={Ma, Xinyin and Wan, Guangnian and Yu, Runpeng and Fang, Gongfan and Wang, Xinchao},
  journal={arXiv preprint arXiv:2502.09601},
  year={2025}
}

@inproceedings{rein2024gpqa,
  title={Gpqa: A graduate-level google-proof q\&a benchmark},
  author={Rein, David and Hou, Betty Li and Stickland, Asa Cooper and Petty, Jackson and Pang, Richard Yuanzhe and Dirani, Julien and Michael, Julian and Bowman, Samuel R},
  booktitle={First Conference on Language Modeling},
  year={2024}
}

@inproceedings{lightman2023let,
  title     = {{Let's Verify Step by Step}},
  author    = {Lightman, Hunter and Kosaraju, Vineet and Burda, Yuri and Edwards, Harrison and Baker, Bowen and Lee, Teddy and Leike, Jan and Schulman, John and Sutskever, Ilya and Cobbe, Karl},
  booktitle = {International Conference on Learning Representations},
  year      = {2024},
  url       = {https://mlanthology.org/iclr/2024/lightman2024iclr-let/}
}

@article{wu2025unlocking,
  title={Unlocking efficient long-to-short llm reasoning with model merging},
  author={Wu, Han and Yao, Yuxuan and Liu, Shuqi and Liu, Zehua and Fu, Xiaojin and Han, Xiongwei and Li, Xing and Zhen, Hui-Ling and Zhong, Tao and Yuan, Mingxuan},
  journal={arXiv preprint arXiv:2503.20641},
  year={2025}
}

@misc{shao2024deepseekmath,
      title={DeepSeekMath: Pushing the Limits of Mathematical Reasoning in Open Language Models}, 
      author={Zhihong Shao and Peiyi Wang and Qihao Zhu and Runxin Xu and Junxiao Song and Xiao Bi and Haowei Zhang and Mingchuan Zhang and Y. K. Li and Y. Wu and Daya Guo},
      year={2024},
      eprint={2402.03300},
      archivePrefix={arXiv},
      primaryClass={cs.CL},
      url={https://arxiv.org/abs/2402.03300}, 
}

@misc{jain2025livecodebench,
      title={LiveCodeBench: Holistic and Contamination Free Evaluation of Large Language Models for Code}, 
      author={Naman Jain and King Han and Alex Gu and Wen-Ding Li and Fanjia Yan and Tianjun Zhang and Sida Wang and Armando Solar-Lezama and Koushik Sen and Ion Stoica},
      year={2024},
      eprint={2403.07974},
      archivePrefix={arXiv},
      primaryClass={cs.SE},
      url={https://arxiv.org/abs/2403.07974}, 
}

@inproceedings{yu2026dapo,
 author = {Yu, Qiying and Zhang, Zheng and Zhu, Ruofei and Yuan, Yufeng and Zuo, Xiaochen and Yue, Yu and Dai, Weinan and Fan, Tiantian and Liu, Gaohong and liu, juncai and Liu, LingJun and Liu, Xin and Lin, Haibin and Lin, Zhiqi and Ma, Bole and Sheng, Guangming and Tong, Yuxuan and Zhang, Chi and Zhang, Mofan and Zhang, Ru and Zhang, Wang and Zhu, Hang and Zhu, Jinhua and Chen, Jiaze and Chen, Jiangjie and Wang, Chengyi and Yu, Hongli and Song, Yuxuan and Wei, Xiangpeng and Zhou, Hao and Liu, Jingjing and Ma, Wei-Ying and Zhang, Ya-Qin and Yan, Lin and Wu, Yonghui and Wang, Mingxuan},
 booktitle = {Advances in Neural Information Processing Systems},
 doi = {10.52202/085713-3775},
 editor = {D. Belgrave and C. Zhang and H. Lin and R. Pascanu and P. Koniusz and M. Ghassemi and N. Chen},
 pages = {113222--113244},
 publisher = {Curran Associates, Inc.},
 title = {DAPO: An Open-Source LLM Reinforcement Learning System at Scale},
 url = {https://proceedings.neurips.cc/paper_files/paper/2025/file/a4277440d50f1f15d2cb4c14f7e0c0d2-Paper-Conference.pdf},
 volume = {38, Main Conference},
 year = {2025}
}

@inproceedings{wei2022chain,
 author = {Wei, Jason and Wang, Xuezhi and Schuurmans, Dale and Bosma, Maarten and ichter, brian and Xia, Fei and Chi, Ed and Le, Quoc V and Zhou, Denny},
 booktitle = {Advances in Neural Information Processing Systems},
 doi = {10.52202/068431-1800},
 editor = {S. Koyejo and S. Mohamed and A. Agarwal and D. Belgrave and K. Cho and A. Oh},
 pages = {24824--24837},
 publisher = {Curran Associates, Inc.},
 title = {Chain-of-Thought Prompting Elicits Reasoning in Large Language Models},
 url = {https://proceedings.neurips.cc/paper_files/paper/2022/file/9d5609613524ecf4f15af0f7b31abca4-Paper-Conference.pdf},
 volume = {35},
 year = {2022}
}

@article{kojima2022large,
  title={Large language models are zero-shot reasoners},
  author={Kojima, Takeshi and Gu, Shixiang Shane and Reid, Machel and Matsuo, Yutaka and Iwasawa, Yusuke},
  journal={Advances in neural information processing systems},
  volume={35},
  pages={22199--22213},
  year={2022}
}

@misc{sui2025stop,
      title={Stop Overthinking: A Survey on Efficient Reasoning for Large Language Models}, 
      author={Yang Sui and Yu-Neng Chuang and Guanchu Wang and Jiamu Zhang and Tianyi Zhang and Jiayi Yuan and Hongyi Liu and Andrew Wen and Shaochen Zhong and Na Zou and Hanjie Chen and Xia Hu},
      year={2025},
      eprint={2503.16419},
      archivePrefix={arXiv},
      primaryClass={cs.CL},
      url={https://arxiv.org/abs/2503.16419}, 
}

@misc{chen2024not,
      title={Do NOT Think That Much for 2+3=? On the Overthinking of o1-Like LLMs}, 
      author={Xingyu Chen and Jiahao Xu and Tian Liang and Zhiwei He and Jianhui Pang and Dian Yu and Linfeng Song and Qiuzhi Liu and Mengfei Zhou and Zhuosheng Zhang and Rui Wang and Zhaopeng Tu and Haitao Mi and Dong Yu},
      year={2025},
      eprint={2412.21187},
      archivePrefix={arXiv},
      primaryClass={cs.CL},
      url={https://arxiv.org/abs/2412.21187}, 
}

@article{aytes2025sketch,
  title={Sketch-of-thought: Efficient llm reasoning with adaptive cognitive-inspired sketching},
  author={Aytes, Simon A and Baek, Jinheon and Hwang, Sung Ju},
  journal={arXiv preprint arXiv:2503.05179},
  year={2025}
}

@article{su2025between,
  title={Between underthinking and overthinking: An empirical study of reasoning length and correctness in llms},
  author={Su, Jinyan and Healey, Jennifer and Nakov, Preslav and Cardie, Claire},
  journal={arXiv preprint arXiv:2505.00127},
  year={2025}
}

@article{han2024token,
  title={Token-budget-aware llm reasoning},
  author={Han, Tingxu and Wang, Zhenting and Fang, Chunrong and Zhao, Shiyu and Ma, Shiqing and Chen, Zhenyu},
  journal={arXiv preprint arXiv:2412.18547},
  year={2024}
}

@inproceedings{kang2025c3ot,
  title={C3ot: Generating shorter chain-of-thought without compromising effectiveness},
  author={Kang, Yu and Sun, Xianghui and Chen, Liangyu and Zou, Wei},
  booktitle={Proceedings of the AAAI Conference on Artificial Intelligence},
  volume={39},
  number={23},
  pages={24312--24320},
  year={2025}
}

@article{xia2025tokenskip,
  title={Tokenskip: Controllable chain-of-thought compression in llms},
  author={Xia, Heming and Li, Yongqi and Leong, Chak Tou and Wang, Wenjie and Li, Wenjie},
  journal={arXiv preprint arXiv:2502.12067},
  year={2025}
}

@misc{lighteval,
  author = {Habib, Nathan and Fourrier, Clémentine and Kydlíček, Hynek and Wolf, Thomas and Tunstall, Lewis},
  title = {LightEval: A lightweight framework for LLM evaluation},
  year = {2023},
  version = {0.11.0},
  url = {https://github.com/huggingface/lighteval}
}

@misc{luo2025deepscaler,
  title={{DeepScaleR}: Surpassing {O1-Preview} with a {1.5B} Model by Scaling {RL}},
  author={Michael Luo and Sijun Tan and Justin Wong and Xiaoxiang Shi
    and William Y. Tang and Manan Roongta and Colin Cai and Jeffrey Luo
    and Tianjun Zhang and Li Erran Li and Raluca Ada Popa and Ion Stoica},
  year={2025},
  howpublished={\href{https://pretty-radio-b75.notion.site/DeepScaleR-Surpassing-O1-Preview-with-a-1-5B-Model-by-Scaling-RL-19681902c1468005bed8ca303013a4e2}{Notion Blog}}
}

@article{cobbe2021gsm8k,
  title={Training Verifiers to Solve Math Word Problems},
  author={Cobbe, Karl and Kosaraju, Vineet and Bavarian, Mohammad and Chen, Mark and Jun, Heewoo and Kaiser, Lukasz and Plappert, Matthias and Tworek, Jerry and Hilton, Jacob and Nakano, Reiichiro and Hesse, Christopher and Schulman, John},
  journal={arXiv preprint arXiv:2110.14168},
  year={2021}
}

@misc{cuadron2025danger,
      title={The Danger of Overthinking: Examining the Reasoning-Action Dilemma in Agentic Tasks}, 
      author={Alejandro Cuadron and Dacheng Li and Wenjie Ma and Xingyao Wang and Yichuan Wang and Siyuan Zhuang and Shu Liu and Luis Gaspar Schroeder and Tian Xia and Huanzhi Mao and Nicholas Thumiger and Aditya Desai and Ion Stoica and Ana Klimovic and Graham Neubig and Joseph E. Gonzalez},
      year={2025},
      eprint={2502.08235},
      archivePrefix={arXiv},
      primaryClass={cs.AI},
      url={https://arxiv.org/abs/2502.08235}, 
}

@article{hou2025thinkprune,
  title={ThinkPrune: Pruning Long Chain-of-Thought of LLMs via Reinforcement Learning},
  author={Hou, Bairu and Zhang, Yang and Ji, Jiabao and Liu, Yujian and Qian, Kaizhi and Andreas, Jacob and Chang, Shiyu},
  journal={arXiv preprint arXiv:2504.01296},
  year={2025}
}

@misc{yang2025qwen3,
      title={Qwen3 Technical Report}, 
      author={An Yang and Anfeng Li and Baosong Yang and Beichen Zhang and Binyuan Hui and Bo Zheng and Bowen Yu and Chang Gao and Chengen Huang and Chenxu Lv and Chujie Zheng and Dayiheng Liu and Fan Zhou and Fei Huang and Feng Hu and Hao Ge and Haoran Wei and Huan Lin and Jialong Tang and Jian Yang and Jianhong Tu and Jianwei Zhang and Jianxin Yang and Jiaxi Yang and Jing Zhou and Jingren Zhou and Junyang Lin and Kai Dang and Keqin Bao and Kexin Yang and Le Yu and Lianghao Deng and Mei Li and Mingfeng Xue and Mingze Li and Pei Zhang and Peng Wang and Qin Zhu and Rui Men and Ruize Gao and Shixuan Liu and Shuang Luo and Tianhao Li and Tianyi Tang and Wenbiao Yin and Xingzhang Ren and Xinyu Wang and Xinyu Zhang and Xuancheng Ren and Yang Fan and Yang Su and Yichang Zhang and Yinger Zhang and Yu Wan and Yuqiong Liu and Zekun Wang and Zeyu Cui and Zhenru Zhang and Zhipeng Zhou and Zihan Qiu},
      year={2025},
      eprint={2505.09388},
      archivePrefix={arXiv},
      primaryClass={cs.CL},
      url={https://arxiv.org/abs/2505.09388}, 
}

@article{yu2025long,
  title={Long-short chain-of-thought mixture supervised fine-tuning eliciting efficient reasoning in large language models},
  author={Yu, Bin and Yuan, Hang and Li, Haotian and Xu, Xueyin and Wei, Yuliang and Wang, Bailing and Qi, Weizhen and Chen, Kai},
  journal={arXiv preprint arXiv:2505.03469},
  year={2025}
}

@inproceedings{marczak2024magmax,
  title={Magmax: Leveraging model merging for seamless continual learning},
  author={Marczak, Daniel and Twardowski, Bart{\l}omiej and Trzci{\'n}ski, Tomasz and Cygert, Sebastian},
  booktitle={European Conference on Computer Vision},
  pages={379--395},
  year={2024},
  organization={Springer}
}

@article{yang2023adamerging,
  title={Adamerging: Adaptive model merging for multi-task learning},
  author={Yang, Enneng and Wang, Zhenyi and Shen, Li and Liu, Shiwei and Guo, Guibing and Wang, Xingwei and Tao, Dacheng},
  journal={arXiv preprint arXiv:2310.02575},
  year={2023}
}

@article{gangwal2025merge,
  title={Merge Now, Regret Later: The Hidden Cost of Model Merging is Adversarial Transferability},
  author={Gangwal, Ankit and Sharma, Aaryan Ajay},
  journal={arXiv preprint arXiv:2509.23689},
  year={2025}
}

@inproceedings{utans1996weight,
  title={Weight averaging for neural networks and local resampling schemes},
  author={Utans, Joachim},
  booktitle={Proc. AAAI-96 Workshop on Integrating Multiple Learned Models. AAAI Press},
  pages={133--138},
  year={1996},
  organization={Citeseer}
}

@article{ilharco2022editing,
  title={Editing models with task arithmetic},
  author={Ilharco, Gabriel and Ribeiro, Marco Tulio and Wortsman, Mitchell and Gururangan, Suchin and Schmidt, Ludwig and Hajishirzi, Hannaneh and Farhadi, Ali},
  journal={arXiv preprint arXiv:2212.04089},
  year={2022}
}

@article{yang2024model,
  title={Model merging in llms, mllms, and beyond: Methods, theories, applications and opportunities},
  author={Yang, Enneng and Shen, Li and Guo, Guibing and Wang, Xingwei and Cao, Xiaochun and Zhang, Jie and Tao, Dacheng},
  journal={arXiv preprint arXiv:2408.07666},
  year={2024}
}

@misc{team2025kimi,
      title={Kimi k1.5: Scaling Reinforcement Learning with LLMs}, 
      author={Kimi Team and Angang Du and Bofei Gao and Bowei Xing and Changjiu Jiang and Cheng Chen and Cheng Li and Chenjun Xiao and Chenzhuang Du and Chonghua Liao and Chuning Tang and Congcong Wang and Dehao Zhang and Enming Yuan and Enzhe Lu and Fengxiang Tang and Flood Sung and Guangda Wei and Guokun Lai and Haiqing Guo and Han Zhu and Hao Ding and Hao Hu and Hao Yang and Hao Zhang and Haotian Yao and Haotian Zhao and Haoyu Lu and Haoze Li and Haozhen Yu and Hongcheng Gao and Huabin Zheng and Huan Yuan and Jia Chen and Jianhang Guo and Jianlin Su and Jianzhou Wang and Jie Zhao and Jin Zhang and Jingyuan Liu and Junjie Yan and Junyan Wu and Lidong Shi and Ling Ye and Longhui Yu and Mengnan Dong and Neo Zhang and Ningchen Ma and Qiwei Pan and Qucheng Gong and Shaowei Liu and Shengling Ma and Shupeng Wei and Sihan Cao and Siying Huang and Tao Jiang and Weihao Gao and Weimin Xiong and Weiran He and Weixiao Huang and Weixin Xu and Wenhao Wu and Wenyang He and Xianghui Wei and Xianqing Jia and Xingzhe Wu and Xinran Xu and Xinxing Zu and Xinyu Zhou and Xuehai Pan and Y. Charles and Yang Li and Yangyang Hu and Yangyang Liu and Yanru Chen and Yejie Wang and Yibo Liu and Yidao Qin and Yifeng Liu and Ying Yang and Yiping Bao and Yulun Du and Yuxin Wu and Yuzhi Wang and Zaida Zhou and Zhaoji Wang and Zhaowei Li and Zhen Zhu and Zheng Zhang and Zhexu Wang and Zhilin Yang and Zhiqi Huang and Zihao Huang and Ziyao Xu and Zonghan Yang and Zongyu Lin},
      year={2025},
      eprint={2501.12599},
      archivePrefix={arXiv},
      primaryClass={cs.AI},
      url={https://arxiv.org/abs/2501.12599}, 
}

@article{wu2025revisiting,
  title={Revisiting model interpolation for efficient reasoning},
  author={Wu, Taiqiang and Yang, Runming and Liu, Tao and Wang, Jiahao and Wong, Ngai},
  journal={arXiv preprint arXiv:2510.10977},
  year={2025}
}

@misc{slime_github,
  author       = {Zilin Zhu and Chengxing Xie and Xin Lv and slime Contributors},
  title        = {slime: An LLM post-training framework for RL Scaling},
  year         = {2025},
  howpublished = {\url{https://github.com/THUDM/slime}},
  note         = {GitHub repository. Corresponding author: Xin Lv},
  urldate      = {2025-06-19}
}

@inproceedings{shoemake1985animating,
  title = {Animating Rotation with Quaternion Curves},
  author = {Shoemake, Ken},
  booktitle = {Proceedings of the 12th Annual Conference on Computer Graphics and Interactive Techniques},
  series = {SIGGRAPH '85},
  pages = {245--254},
  year = {1985},
  publisher = {Association for Computing Machinery},
  address = {New York, NY, USA},
  isbn = {0897911660},
  doi = {10.1145/325334.325242},
  url = {https://doi.org/10.1145/325334.325242}
}

@inproceedings{yadav2023ties,
  title = {TIES-Merging: Resolving Interference When Merging Models},
  author = {Yadav, Prateek and Tam, Derek and Choshen, Leshem and Raffel, Colin and Bansal, Mohit},
  booktitle = {Advances in Neural Information Processing Systems},
  volume = {36},
  pages = {7093--7115},
  year = {2023},
  publisher = {Curran Associates, Inc.},
  url = {https://proceedings.neurips.cc/paper_files/paper/2023/file/1644c9af28ab7916874f6fd6228a9bcf-Paper-Conference.pdf}
}

@inproceedings{yu2024language,
  title = {Language Models are Super Mario: Absorbing Abilities from Homologous Models as a Free Lunch},
  author = {Yu, Le and Yu, Bowen and Yu, Haiyang and Huang, Fei and Li, Yongbin},
  booktitle = {Proceedings of the 41st International Conference on Machine Learning},
  pages = {57755--57775},
  year = {2024},
  editor = {Salakhutdinov, Ruslan and Kolter, Zico and Heller, Katherine and Weller, Adrian and Oliver, Nuria and Scarlett, Jonathan and Berkenkamp, Felix},
  volume = {235},
  series = {Proceedings of Machine Learning Research},
  month = {21--27 Jul},
  publisher = {PMLR},
  url = {https://proceedings.mlr.press/v235/yu24p.html}
}

@article{deep2024della,
  title = {DELLA-Merging: Reducing Interference in Model Merging through Magnitude-Based Sampling},
  author = {{Pala Tej Deep} and Bhardwaj, Rishabh and Poria, Soujanya},
  journal = {arXiv preprint arXiv:2406.11617},
  year = {2024},
  archivePrefix = {arXiv},
  eprint = {2406.11617},
  primaryClass = {cs.CL},
  url = {https://arxiv.org/abs/2406.11617}
}

@article{hansen2001completely,
  title = {Completely Derandomized Self-Adaptation in Evolution Strategies},
  author = {Hansen, Nikolaus and Ostermeier, Andreas},
  journal = {Evolutionary Computation},
  volume = {9},
  number = {2},
  pages = {159--195},
  year = {2001},
  publisher = {MIT Press},
  doi = {10.1162/106365601750190398},
  url = {https://doi.org/10.1162/106365601750190398}
}

@inproceedings{wu2026beyond,
  title={Beyond Token Length: Step Pruner for Efficient and Accurate Reasoning in Large Language Models},
  author={Wu, Canhui and Cao, Qiong and Li, Chang and Wang, Zhenfang and Xue, Chao and Fan, Yuwei and Xi, Wei and He, Xiaodong},
  booktitle={Findings of the Association for Computational Linguistics: ACL 2026},
  pages={1953--1974},
  year={2026}
}

\clearpage

\appendix

\section{Appendix}

\subsection{Experimental environment}
We trained on a node with 8$\times$NVIDIA H200 GPUs and Intel Xeon Platinum 8558 CPUs. The total training time was approximately 42 hours.

\subsection{Qwen3-4B training curves}
The training dynamics show that GRIP gradually shortens responses while maintaining a stable reward signal, indicating that the learned interpolation can improve efficiency without collapsing task performance.

\begin{figure}[H]
    \centering
    \includegraphics[width=\columnwidth]{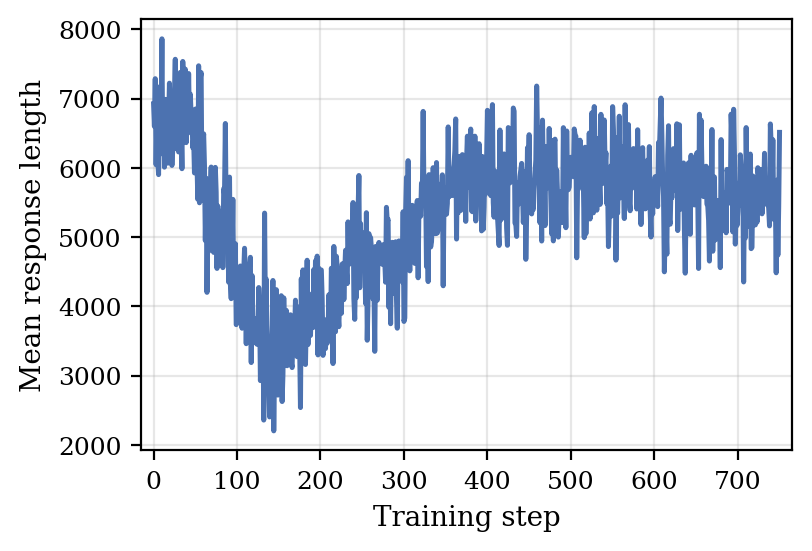}
    \caption{Mean response length during GRIP training on the Qwen3-4B pair through step $750$.}
    \label{fig:qwen4b_response_len}
\end{figure}

\begin{figure}[H]
    \centering
    \includegraphics[width=\columnwidth]{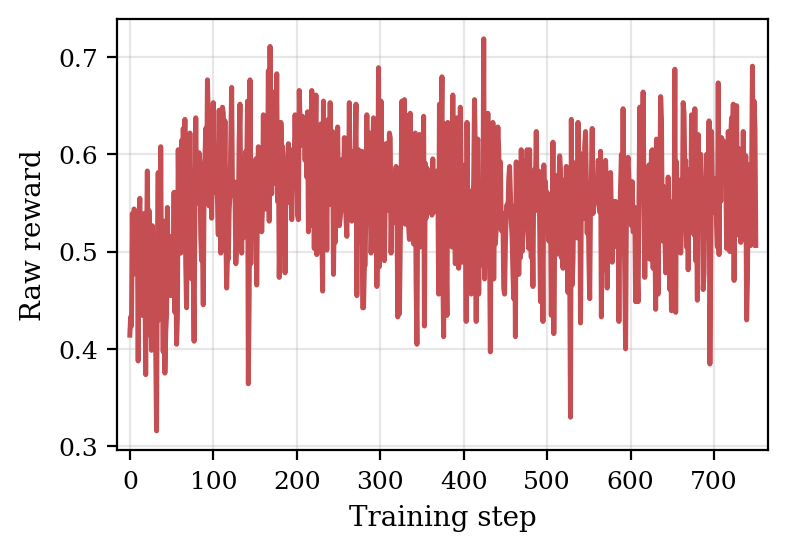}
    \caption{Raw reward during GRIP training on the Qwen3-4B pair through step $750$.}
    \label{fig:qwen4b_reward}
\end{figure}

\subsection{Ablation: layer-wise versus module-wise interpolation}
We compare the module-wise parameterization used by GRIP with a coarser layer-wise alternative. The layer-wise variant assigns one shared coefficient to each transformer layer, together with two special coefficients: one for the final RMSNorm and one shared by the tied input embedding and LM head, resulting in $36+2$ trainable parameters. In contrast, our module-wise design assigns separate coefficients to attention and FFN in every layer, plus the same two special coefficients, resulting in $72+2$ trainable parameters. This distinction matters because Section~\ref{sec:module_sweeps} shows that attention and FFN affect accuracy and response length differently; tying them inside each layer forces a single coefficient to control two modules with different roles.

\begin{table}[H]
\centering
\small
\begin{tabular}{lrr}
\toprule
\textbf{Method} & \textbf{Avg Acc.} & \textbf{Avg Tok.} \\
\midrule
Layer-wise ($36+2$ params) & 73.5 & 7571 \\
Module-wise ($72+2$ params) & \textbf{76.5} & 7930 \\
\bottomrule
\end{tabular}
\caption{Ablation comparing layer-wise and module-wise interpolation on the same five evaluation benchmarks. The layer-wise result is evaluated at step $230$, while the module-wise result is the GRIP configuration reported in Table~\ref{tab:main}. Module-wise interpolation improves average accuracy by $3.0$ points while using a comparable number of generated tokens.}
\label{tab:layerwise_ablation}
\end{table}

The result supports using module-wise rather than layer-wise interpolation. Although the layer-wise model is slightly shorter on average, it loses substantial accuracy because it cannot independently preserve reasoning-sensitive FFN components while allowing attention modules to move differently. The additional parameters in the module-wise design are therefore not merely extra capacity; they encode the structural asymmetry between attention and FFN observed in Figure~\ref{fig:sweep_attn_ffn}.

\subsection{Artifact licenses and terms}
We use publicly available models, frameworks, and evaluation artifacts in accordance with their released licenses and terms. LightEval, LiveCodeBench, GPQA, GSM8K, and MATH-500 are released under the MIT License. SLIME, Qwen3-4B, and AIME25 are released under the Apache-2.0 License. Our use of these artifacts is limited to training and evaluation in the experimental setting described in this paper, and we do not redistribute modified versions of the original datasets or model checkpoints.

\subsection{Data statistics and splits}
GRIP is trained on the DeepScaleR-preview math prompt set used by SLIME, containing $40{,}196$ training examples in JSONL format. We do not construct additional development or test splits from this training set; all reported generalization results use external benchmarks. Evaluation is conducted through LightEval on the official benchmark splits: AIME25 contains $30$ competition problems, MATH500 contains $500$ math problems, GSM8K contains $1{,}319$ grade-school math test problems, GPQA-Diamond contains $198$ graduate-level science questions, and LiveCodeBench code generation v6 contains $175$ programming problems. Table~\ref{tab:main} reports results on these five evaluation sets, and Section~\ref{sec:module_sweeps} uses the same evaluation protocol for coefficient sweeps.

\subsection{Software package versions}
The environment uses Python 3.12.3, CUDA 12.9, SLIME 0.2.4, Megatron-Core 0.16.0rc0, SGLang 0.5.10.post1, Ray 2.55.1, PyTorch 2.9.1+cu129, Transformers 5.3.0, Safetensors 0.7.0, SymPy 1.14.0, NumPy 1.26.4, and Weights \& Biases 0.26.1. Evaluation uses the LightEval codebase at commit \texttt{33acf35f}. We will release the source code for GRIP to support reproducibility.

\subsection{Human subjects and ethics review}
This work did not involve human subjects, crowdworkers, or human annotators. We used only publicly available datasets, models, and automated evaluation pipelines. Therefore, no new data collection protocol involving human participants was conducted, and ethics review board approval was not applicable.

\end{document}